\documentclass[twocolumn,10pt,twosided]{article}

\usepackage[ruled]{algorithm2e} 

\usepackage{amsmath,amssymb}
\usepackage{graphicx}
\usepackage{color}
\usepackage{cite}
\usepackage{multirow}
\usepackage{soul}

\newcommand{\OO}[1]{\mathcal{O}\left(#1\right)}

\DeclareMathOperator*{\argmin}{\arg\!\min}
\title{Learning event representations\\ for temporal segmentation of image sequences\\ by dynamic graph embedding}

\author{Mariella Dimiccoli 
and Herwig Wendt%
\\
\thanks{Manuscript received August 10, 2019; revised June 19, 2020; October 30, 2020, December 4, 2020, December 8, 2020; accepted December 9, 2020.}
\thanks{M. Dimiccoli is with the Institut de Rob\`otica i Inform\`atica Industrial, CSIC-UPC, Barcelona, Spain ({\tt mdimiccoli@iri.upc.edu}).}%
\thanks{H. Wendt is with the Institut de Recherche en Informatique de Toulouse, CNRS, University of Toulouse,  France ({\tt herwig.wendt@irit.fr}).}%
\thanks{This work was partially supported by the Spanish Ministry of Economy and Competitiveness and the European Regional Development Fund (MINECO/ERDF, EU) through the program Ramon y Cajal and the national Spanish projects PID2019-110977GA-I00, RED2018-102511-T and 2017 SGR 1785. We acknowledge the support of NVIDIA Corporation with the donation of Titan Xp GPUs.}
}

\date{December 8, 2020}

\begin{document}

\maketitle

\begin{abstract}
\boldmath
Recently, self-supervised learning has proved to be effective to learn representations of \textit{events} suitable for temporal segmentation in image sequences, where events are understood as sets of temporally adjacent images that are semantically perceived as a whole. However, although this approach does not require expensive manual annotations,  it is data hungry and suffers from domain adaptation problems. As an alternative, in this work, we propose a novel approach for learning event representations named \textit{Dynamic Graph Embedding} (DGE). The assumption underlying our model is that a sequence of images can be represented by a graph that encodes both semantic and temporal similarity. The key novelty of DGE is to learn jointly the graph and its graph embedding. At its core, DGE  works by iterating over two steps:  1) updating the graph representing the semantic and temporal similarity of the data based on the current data representation, and 2) updating the data representation to take into account the current data graph structure. The main advantage of DGE over state-of-the-art self-supervised approaches is that it does not require any training set, but instead learns iteratively from the data itself a low-dimensional embedding that reflects their temporal and semantic similarity. Experimental results on two benchmark datasets of real image sequences captured at regular time intervals demonstrate that the proposed DGE leads to event representations effective for temporal segmentation. In particular, it achieves robust temporal segmentation on the EDUBSeg and EDUBSeg-Desc benchmark datasets, outperforming the state of the art. Additional experiments on two Human Motion Segmentation benchmark datasets demonstrate the generalization capabilities of the proposed DGE.
\end{abstract}

\section{Introduction}

\begin{figure}[!t]
\centering
\includegraphics[width=\columnwidth]{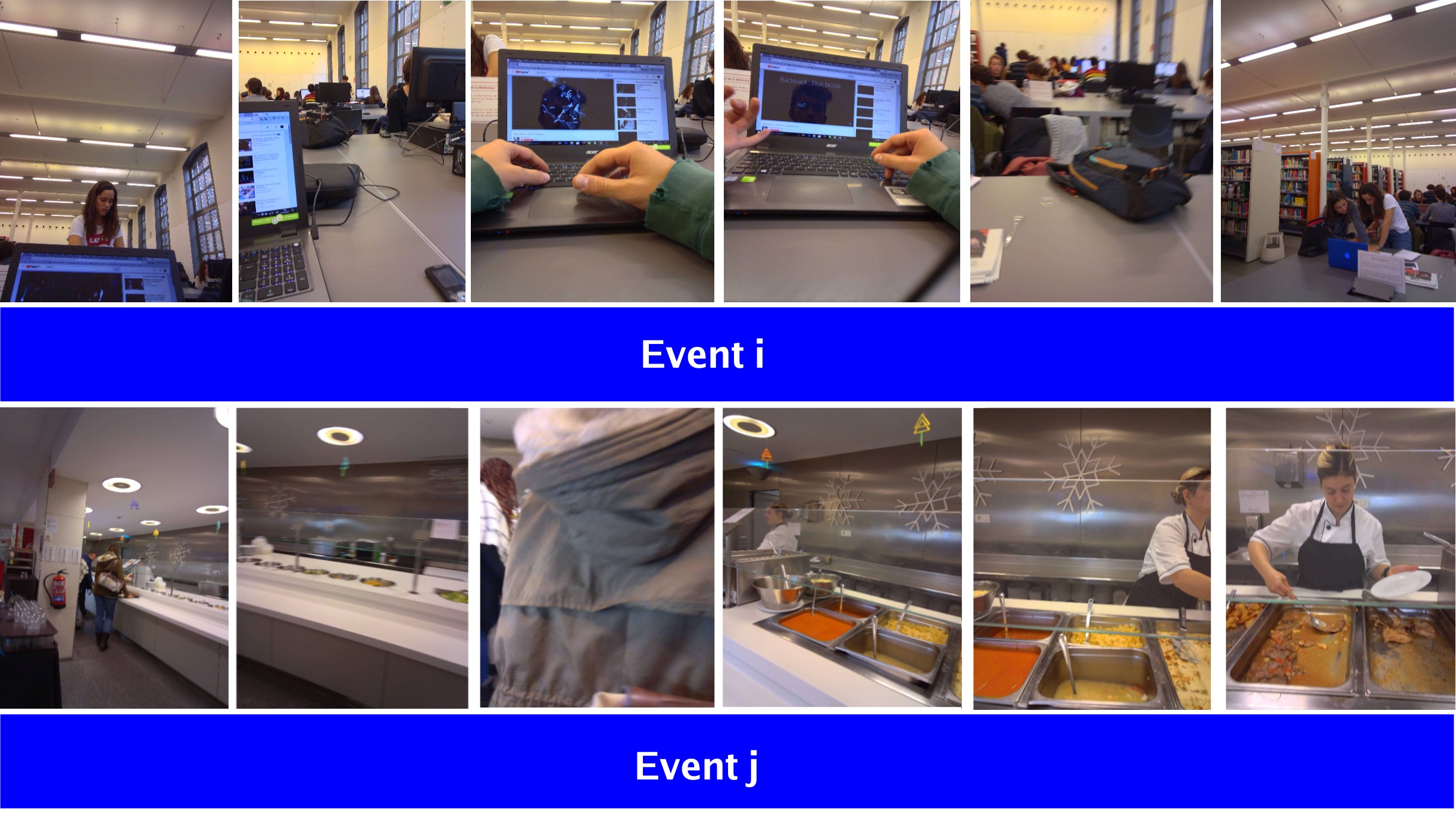}
\caption[]{
Temporally adjacent frames in two events in first-person image sequences.}
\label{fig:difficult_events}
\end{figure}

Temporal segmentation of videos and image sequences has a long story of research since it is crucial not only to video understanding but also to video browsing, indexing and summarization \cite{koprinska2001temporal,money2008video,del2017summarization}. With the proliferation of wearable cameras in recent years, the field is facing new challenges. Indeed, wearable cameras allow to capture, from a first-person (egocentric) perspective, and ``in the wild'', long unconstrained videos ($\approx$35fps) and image sequences (aka \emph{photostreams}, $\approx$2fpm). Due to their low temporal resolution, the segmentation of first-person image sequences is particularly challenging, and has received special attention from the community \cite{lin2006structuring,doherty2008combining,doherty2008investigating,talavera2015r,paci2016context,lin2017vci2r,yamamoto2017pbg,dimiccoli2017sr,del2018predicting,dias2019learning,dimiccoli2019enhancing}. Indeed, abrupt changes in appearance may arise even between temporally adjacent frames within an event due to sudden camera movements and the low frame rate, making it difficult to distinguish them from event transitions. While for a human observer it is relatively easy to segment egocentric image sequences into discrete units, this poses serious difficulties for automated temporal segmentation (see Figure~\ref{fig:difficult_events} for an illustration). In particular, classical spatio-temporal video representations, that typically rely on motion information \cite{palou2013hierarchical,tran2015learning,wang2019self},  cannot be reliably computed on photostreams due to this lack of temporal continuity \cite{bolanos2017toward}.

Given the limited amount of annotated data,
current state-of-the-art approaches for the temporal segmentation of first-person image sequences aim at obtaining event representations by encoding the temporal context of each frame in an unsupervised fashion \cite{dias2019learning,del2018predicting}. These methods rely on neural or recurrent neural networks and are generally based on the idea of learning event representations by training the network to predict past and future frame representations. Recurrent neural networks have proved to be more efficient than simple neural networks for the temporal prediction task. The main limitation of these approaches is that they must rely on large training datasets to yield state-of-the-art performance. Even if, in this case, training data do not require manual annotations, they can nevertheless introduce a bias and the learnt models can suffer from the domain adaptation problem. For instance, in the case of temporal segmentation of image sequences, the models will be difficult to generalize to data acquired with a camera with different field of view or for people having different lifestyles. 

\begin{figure}[!t]
\centering
\includegraphics[width=\columnwidth]{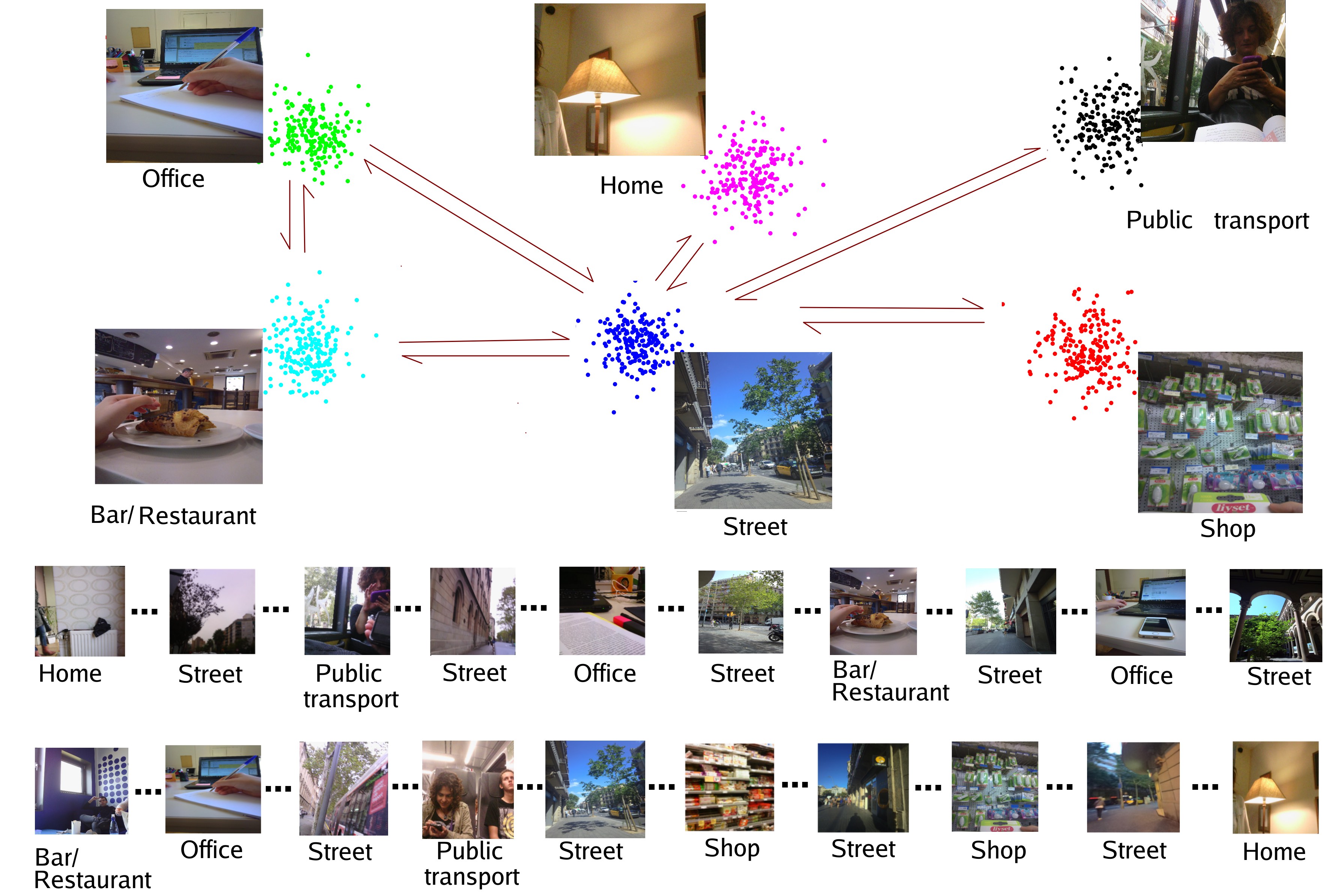}
\caption[]{
The assumption underlying our learning approach is that image sequences captured at regular time intervals (2fpm) can be organized into a graph structure, where each community in the graph corresponds to a particular semantic context. 
Points of the same color in the figure are related semantically and may be more or less related at temporal level. The arrows indicate temporal transitions between communities. They have only a visualization purpose, since temporal transition are between pairs of points.
}
\label{fig:main_idea}
\end{figure}

In this paper, we aim at overcoming this limitation with a novel approach that is able to unveil a representation that encodes the temporal and semantic similarity of an image sequence from the single sequence itself. With this goal in mind, we propose to learn event representations as an embedding on a graph. 
Our model is based on the assumption that each event belongs to a particular semantic context that can be shared across semantically similar events. These semantic contexts can be represented as communities on an unknown underlying graph. 
In particular, graph nodes correspond to individual frames, and edges between nodes encode frame similarity. The communities are understood here as sets of nodes (frames) that are interconnected by edges with large weights.
Moreover, the graph weights reflect not only temporal proximity, but also semantic similarity, which is understood here as similarity in terms of high-level visual features. This is motivated by neuroscientific findings which show that neural representations of events arise from temporal community structures \cite{schapiro2013neural} and suggest that frames which share the context are grouped together in the representational space. In Figure~\ref{fig:main_idea} we illustrate this idea by means of an egocentric image sequence capturing the full day of a person: going from \textit{home} to \textit{work} using \textit{public transports}, having a lunch break in a \textit{restaurant} and going back to \textit{home} after doing some \textit{shopping}, etc.  Each point cloud corresponds with images similar in appearance and most of them are visited multiple times. This means that every pair of images in a point cloud is related semantically, but they could or could not be related at temporal level.

Based on this model, the proposed solution consists in learning simultaneously the graph structure (encoded by its weights) and the data representation. This is achieved by iterating over two alternate steps: 1) update of the graph structure as a function of the current data representation, where the graph structure is assumed to encode a finite number of communities, and 2) update of the data representation as a function of the current graph structure in a low-dimensional embedding space. We term this solution \emph{dynamic graph embedding} (DGE). We provide illustrative experiments on synthetic data, and we validate the proposed approach on two real world benchmark datasets for first-person image sequence temporal segmentation. Our framework is the first attempt to learn simultaneously graph structure and data representations for temporal segmentation of image sequences. 

Our main contributions are: (i) we re-frame the event learning problem as the problem of learning a graph embedding, (ii) we introduce an original graph initialization approach based on the concept of temporal self-similarity, (iii) we propose a novel technical approach to solve the graph embedding problem when the underlying graph structure is unknown, (iv) we demonstrate that the learnt graph embedding  is suitable for the task of temporal segmentation, achieving state-of-the-art results on 
two challenging reference benchmark datasets \cite{dimiccoli2017sr,bolanos2018egocentric}, without relying on any training set for learning the representation,
(v) we show that the proposed DGE generalizes to other problems, yielding state-of-the-art results also on 
two reference benchmark datasets for the Human Motion Segmentation problem.

The structure of the paper is as follows. Section \ref{sec:SoA} highlights related work on data representation learning on graphs and on the temporal segmentation of videos and image sequences. In Section \ref{sec:formulation} we introduce our problem formulation while in Sections \ref{subsec:init} to \ref{sec:synthexp} we detail the proposed graph embedding model. The performance of our algorithm on real world data are evaluated in Section \ref{sec:exp}. In Section \ref{sec:conclusions}, we conclude on our contributions and results.

\section{Related work}
\label{sec:SoA}
\subsection{Geometric learning}
\label{sec:geoLearn}

The proposed approach lies in the field of geometric learning, which is an umbrella term for those techniques that work in non-Euclidean domains such as graphs and manifolds. Following \cite{bronstein2017geometric}, geometric learning approaches either deal with analyzing functions defined on a given non-Euclidean domain or address the problem of characterizing the structure of the data. The former case includes  methods dealing with manifolds \cite{boscaini2016learning,masci2015geodesic} as well as methods of signal processing on graphs \cite{ortega2018graph}. These allow to generalize CNN to graphs \cite{niepert2016learning,defferrard2016convolutional} by defining an operation similar to convolution in the graph spectral domain.
In the latter case, which is more closely related with the method proposed in this paper, the goal is to learn an embedding of the data in a low-dimensional space such that the geometric relations in the embedding space reflect the graph structure. These methods are commonly referred to as \textit{node-embedding} and can be understood  from an encoder-decoder perspective \cite{hamilton2017representation}. Given a graph $\mathcal{G}= (\mathcal{V},\mathcal{E})$, where $\mathcal{V}$ and $\mathcal{E}$ represent the set of nodes and edges of the graph respectively, the encoder maps each node $v \in \mathcal{V}$ of $\mathcal{G}$ in a low-dimensional space. The decoder is a function defined in the embedding space that acts  on node pairs to compute a similarity $S$ between the nodes. Therefore the graph embedding problem can be formulated as the problem of optimizing decoder and encoder mappings such that they minimize the discrepancy between similarity values in the embedding and original feature space. 

Within the general encoder-decoder framework, node embedding algorithms can be roughly classified into two classes. The first class covers shallow embedding methods, including matrix factorization  \cite{ahmed2013distributed,ou2016asymmetric,cao2015grarep} and random-walk based approaches \cite{perozzi2014deepwalk,grover2016node2vec,tang2015line,chen2018harp}. The second class subsumes generalized encoder-decoder architectures \cite{cao2016deep,wang2016structural,hinton2006reducing,hamilton2017inductive}. In shallow embedding approaches, the encoder function acts simply as a lookup function and the input nodes $v_i \in \mathcal{V}$  are represented as one-hot vectors, so that they cannot leverage node attributes during encoding. 
Instead, in generalized encoder-decoder architectures \cite{cao2016deep,wang2016structural,hinton2006reducing} the encoders depend on the structure and attributes of the graph. In particular, convolutional encoders \cite{hamilton2017inductive} rely on node features to generate embeddings for a node by aggregating information from its local neighborhood, in a manner similar to the receptive field of a convolutional kernel in image processing. As the process is iterated, the node embedding contains information aggregated from further and further reaches of the graph. Closely related to convolutional encoders are Graph Neural Networks (GNNs). The main difference is that GNNs capture the graph's internal dependencies via message passing between its nodes. Moreover, GNNs can utilize  node  attributes  and  node  labels to  train  model  parameters  end-to-end  for  a  specific  task in a semi-supervised fashion \cite{henaff2015deep,garcia2017few,defferrard2016convolutional,niepert2016learning}.

In all these methods, the graph structure is assumed to be given by the problem domain. For instance, the graph structure for social networks can be inferred from the connections between people. However, in the case of temporal segmentation considered here the problem is non-structural since the graph structure is not given by the problem domain. Instead, it needs to be determined together with the node embedding.

\subsection{Event segmentation}

Extensive research has been conducted to temporally segment videos and image sequences into events.  Early approaches aimed at segmenting edited videos such as TV programs and movies \cite{yeung1998segmentation,yuan2007formal,chasanis2009movie,liu2011exploiting,liu2013learning} into commercial, news-related or movie events. 
This includes the use of the concept of Logical Story Units (LSU), defined as \textit{a series of shots that communicate a unified action with a common setting and time}; in particular,
\cite{yeung1998segmentation} proposed a method to segment TV programs into LSUs by firstly clustering given video shots and then building a Scene Transition Graph with nodes corresponding to the clusters and edges to temporal transitions.
More recently, with the advent of wearable cameras and camera equipped smartphones, there has been an increasing interest in segmenting untrimmed videos or image sequences captured by nonprofessionals into semantically homogeneous units \cite{hu2011survey,tang2012learning,Iwan2017}. In particular, videos or image sequences captured by a wearable camera are typically long and unconstrained  \cite{bolanos2017toward}. Therefore it is important for the user to have them segmented into semantically meaningful chapters. In addition to appearance-based features \cite{bettadapura2016discovering,xu2015gaze}, motion features have been extensively used for temporal segmentation of both third-person videos \cite{koprinska2001temporal,tang2012learning} and first-person videos \cite{huang2017egocentric,spriggs2009temporal,lee2015predicting,poleg2014temporal}. In \cite{spriggs2009temporal}, motion cues from a wearable inertial sensor are leveraged for the temporal segmentation of human motion into actions. Lee and Grauman \cite{lee2015predicting} used temporally constrained clustering of motion and visual features to determine whether the differences in appearance correspond to event boundaries or just to abrupt head movements. Poleg et al. \cite{poleg2014temporal} proposed to use integrated motion vectors to segment egocentric videos into a hierarchy of long-term activities whose first level corresponds to static/transit activities.

However, motion information is not available in first-person image sequences that are the main focus of this paper. In addition, given the limited amount of annotated data, event segmentation is very often performed by using a clustering approach that takes as input hand-crafted visual features such as color \cite{lin2006structuring}, MPEG7 descriptors \cite{doherty2008investigating},  a combination of environmental sensor data, SIFT, SURF and MPEG7 descriptors \cite{doherty2008combining}, or a combination of CNN-based features \cite{lin2017vci2r,yamamoto2017pbg}. Tavalera et al.~\cite{talavera2015r} proposed to combine agglomerative clustering with a change detection method within a graph-cut energy minimization framework. Later on, \cite{dimiccoli2017sr} extended this framework and proposed an improved feature representation by building a vocabulary of concepts.
Paci et al.~\cite{paci2016context} proposed a Siamese ConvNets based approach that aims at learning a similarity function between low temporal resolution egocentric images. Recently, \cite{dias2019learning} proposed to learn event representations as the byproduct of learning to predict the temporal context. In this work, the single image sequence itself is employed to learn the new representation by using an autoencoder model or a LSTM encoder-decoder model, without relying on a training dataset. Molino et al.~\cite{del2018predicting} later proposed a similar LSTM based model that achieved impressive results on the EDUBSeg dataset by leveraging on more powerful initial features and by relying on a large training dataset (over 1.2 million images).

Here, we propose a new model that as in \cite{dias2019learning} does not make use of any training set for learning the temporal event representation, but achieves state-of-the-art results. In particular, it outperforms \cite{del2018predicting} on the EDUBSeg and EDUBSeg-Desc benchmarks \cite{dimiccoli2017sr,bolanos2018egocentric}.

\setlength{\tabcolsep}{2pt}
\begin{figure*}[!t]
   \centering
\includegraphics[width=0.85\linewidth]{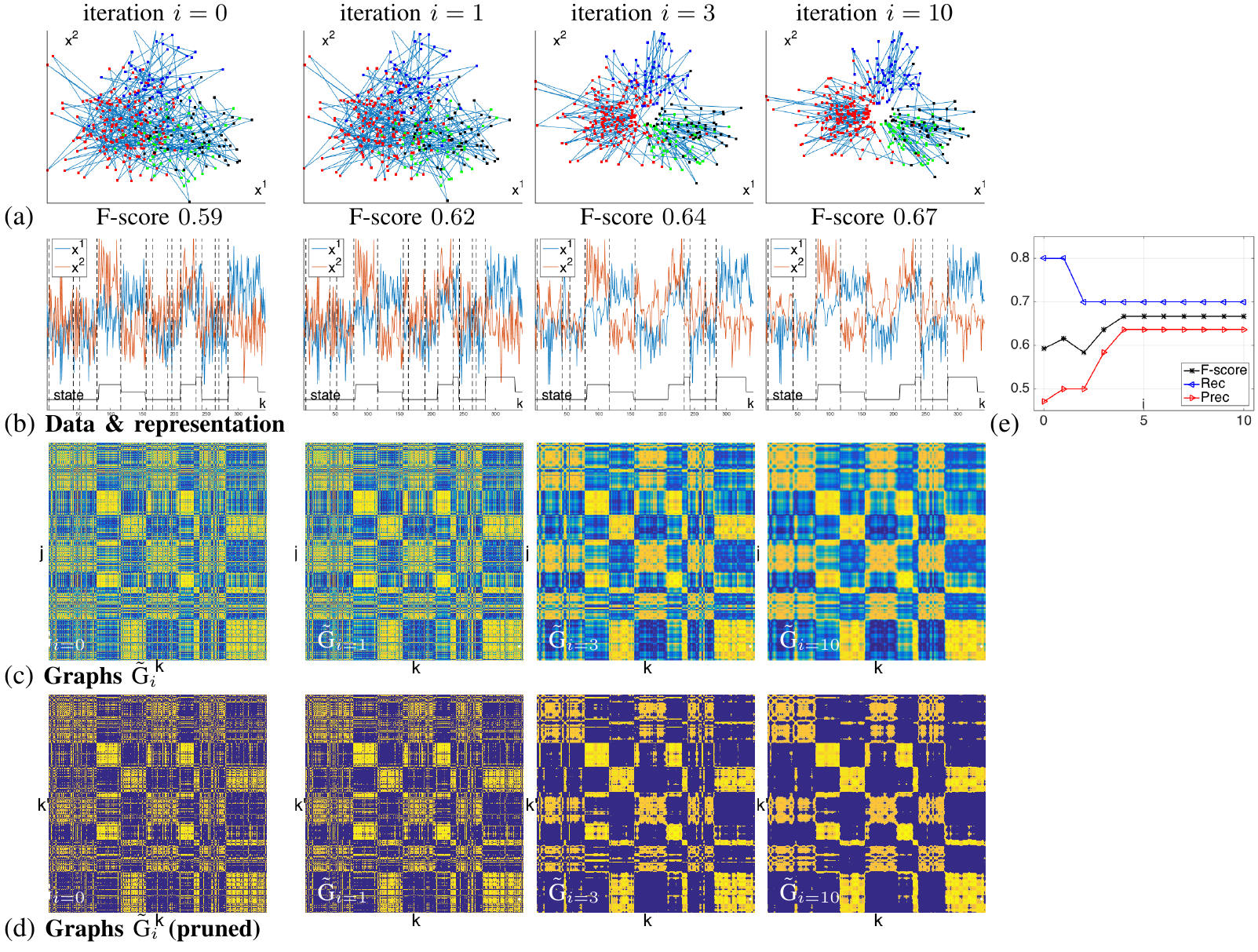}\vspace{-3mm}
\caption[]{%
{\bf\boldmath Illustration of DGE on synthetic data modeling $4$ communities in $d=2$ dimensions}: (a) scatter plot of features $\tilde X_i$ for iterations $i=(0,1,3,10)$ (left to right column, respectively); color indicates community ground truth, solid lines indicate temporal adjacency; (b) the features $\tilde X_i$ plotted as time series, with estimated segment boundaries (vertical dashed lines) and community ground truth (solid black); (c) the corresponding graphs $\mathcal{\tilde{G}}_i$ and (d) $\mathcal{\tilde{G}}_i$ after removal of edges with weight smaller than 0.7 times that of the strongest edge (yellow and blue color correspond with strong and weak similarity, respectively); (e) F-score, precision and recall for estimated segmentation as a function of iteration number $i$.
}
\label{fig:synt}
\end{figure*}

\section{Dynamic Graph Embedding (DGE)}
\label{sec:DGE}

\subsection{Problem formulation and proposed model}
\label{sec:formulation}

We formulate the event learning problem as a geometric learning problem. More specifically, given a set of data points (the frames of the image sequence) embedded into a high-dimensional Euclidean space (the initial data representation), we assume that these data points are organized in a graph in an underlying low-dimensional space. Here, graph nodes correspond to individual frames, and edges between nodes encode frame similarity.
Our \textit{a priori} on the structure of the graph is that it consists of a finite number of communities (sets of nodes (frames) that are interconnected by edges with large weights) corresponding to different semantic contexts. 
Since  along an image sequence a same community can be visited several times at different time intervals, we assume that edges between nodes belonging to different communities correspond to transitions between different semantic contexts. In contrast, edges between nodes belonging to the same community correspond to transitions between nodes sharing the same semantic context, being them temporally adjacent or not. This structure implicitly assumes that the graph topology \emph{models jointly temporal and semantic similarity relations}. 

More formally, let $X \in \mathbb{R}^N \times \mathbb{R}^n$ denote the $n$-dimensional feature vectors
for a given sequence of $N$ images, and $S$ a similarity kernel. We aim at finding a fully connected, weighted graph $\mathcal{\tilde{G}} =(\tilde{X},\mathcal{E},\mathcal{\tilde{W}})$ with node embedding $\tilde{X} \in \mathbb{R}^N \times \mathbb{R}^d$  in a low-dimensional space, $d \ll n$, and edge weights $\mathcal{\tilde{W}}$ given by the entries of an affinity matrix $\tilde{\textnormal{G}}=S(\tilde X)$ such that the similarity $\tilde{\textnormal{G}}_{kj}$ between any pair $k,j$ of nodes of the graph reflects both semantic relatedness and temporal adjacency between the images. Semantic relatedness is captured by a similarity function between high-level visual image descriptors, whereas temporal adjacency is imposed through temporal constraints on the edge weights. The above constraints lead to easily grouping the graph nodes in a finite number of communities that each correspond to a different semantic context. As seen in the previous section, in classical node embedding the low-dimensional representation of each node encodes information about the position and the structure of the local neighborhood in the graph. Since all these methods incorporate graph structure in some way, the construction of the underlying graph is extremely important but relatively little explored. In our problem at hand the graph structure is initially unknown since it arises from unknown events. Therefore, we aim at learning jointly the structure of the underlying graph and the node embedding. 

\begin{algorithm}[tb]
\small
\caption{Dynamic Graph Embedding (DGE)}
\label{alg:algo}
\SetKwData{Left}{left}\SetKwData{This}{this}\SetKwData{Up}{up}
\SetKwFunction{Union}{Union}\SetKwFunction{FindCompress}{FindCompress}
\vspace{1mm}
 $\qquad\qquad N\; - \;$ \it length of the image sequence\\
$\qquad\qquad n\;\; -\; $ \it original feature dimension\\
$\qquad\qquad d\;\; -\;$ \it embedding feature dimension ($d\ll n$)\vspace{1mm}\\
\SetKwInOut{Input}{Input}\SetKwInOut{Output}{Output}
\Input{ $\;X\in\mathbb{R}^N \times \mathbb{R}^n$\hfill\it initial feature matrix}
\Output{ $\;\tilde{X}\in\mathbb{R}^N \times \mathbb{R}^d$\hfill\it graph embedded feature matrix}
\BlankLine
\tcc{ \textbf{Graph initialization} \hfill \rm Eqs.~(\ref{equ:NLsim}-\ref{equ:cosdist})}
$\;\;\hat{X}_0=\textnormal{NLmeans}^{1D}(X)\in\mathbb{R}^N \times \mathbb{R}^n$
\hfill{\it denoise initial features}\\%
$\;\; \textnormal{G}_0 = S_{\hat l}(\hat{X}_0) \in \mathbb{R}^N \times \mathbb{R}^N $
\hfill{\it initialize graph in original space}\vspace{1mm}\\%
\tcc{ \textbf{Graph embedding initialization} \hfill\rm Eqs.~(\ref{equ:cosdist}-\ref{equ:init})}
$\;\;\tilde{X}=\textnormal{PCA}_d(\hat{X}_0)\in\mathbb{R}^N \times \mathbb{R}^d$%
\\%
$\;\;\tilde{X}_0=   \argmin_{\tilde{X}} \mathcal{L} (S_{\tilde{l}}(\tilde{X}),\textnormal{G}_0) $%
\hfill{\it initialize embedding features}\\%
$\;\;\tilde{\textnormal{G}}_0 = S_{\tilde l}(\tilde{X}_0) \in \mathbb{R}^N \!\times\! \mathbb{R}^N$%
\hfill{\it initialize graph in embedding space}%
\vspace{2mm}%

\tcc{\textbf{DGE core loop} \hfill \rm Eqs.~(\ref{eq:embedding}-\ref{equ:graph2})}
\For{$i\leftarrow 1$ \KwTo $K$}{\vspace{1mm}
{\footnotesize\bf --- graph embedding update}\\
$\tilde{X}_i =   \argmin_{\tilde{X}} (1-\alpha ) \mathcal{L}_1 (S_{\tilde l}(\tilde{X}),\tilde{\textnormal{G}}_{i-1})+\alpha\mathcal{L}_2 (S_{\tilde l}(\tilde{X}),\textnormal{G}_0)$\\
{\it\hfill update embedding features given current graph $\mathcal{\tilde{G}}_{i-1}$}\vspace{1mm}\\
{\footnotesize\bf --- graph structure update: temporal prior}\\
$\tilde{\textnormal{G}}_i \gets S_{\tilde l}(\tilde{X}_{i}) \ast \mathcal{K}_{p}$ {\it\hfill local average of weights of graph of $\tilde{X}_i$}\\
$\tilde{\textnormal{G}}_i \gets f_{\eta}(\tilde{\textnormal{G}}_i)$ {\it\hfill strengthen temporally adjacent edges}\vspace{1mm}\\
{\footnotesize\bf --- graph structure update: semantic prior}\\
$\mathcal{C} = \textnormal{kmeans}(\tilde{X}_i; N_{C})$
{\it\hfill estimate semantic communities}\\
$\tilde{\textnormal{G}}_i = g_{\mu}(\tilde{\textnormal{G}}_i , \mathcal{C})$%
{\it\hfill encode semantic similarity in graph}
}
\end{algorithm}

\subsection{Graph initialization by nonlocal self-similarity}
\label{subsec:init}

\noindent{\bf Temporal nonlocal self-similarity.\quad}
{To obtain a first coarse estimate of the graph, we apply a nonlocal self-similarity algorithm in the temporal domain to the initial data $X$ that we normalize to the interval $[-1,1]$ \cite{dimiccoli2019enhancing}. The nonlocal self-similarity filtering creates temporal neighborhoods of frames that are likely to be in the same event. Let $X(k)\in \mathbb{R}^n$ denote the $k$-th row of $X$, that is, the vector of $n$ image features at time $k$, $k=1,\ldots,N$. Further, let $\mathcal{N}_k^M=\{k-M,\ldots,k-1,k+1,\ldots,k+M\}$ and $\mathcal{N}_k^L=\{k-L,\ldots,k-1,k+1,\ldots,k+L\}$ denote the indices of the $2M$ and $2L$ neighboring feature vectors of $X(k)$, respectively, with $L>M$. In analogy with 2D data (images) \cite{buades2005non}, the self-similarity function of  $X(k)$ in a temporal sequence, conditioned to its temporal neighborhood $j\in\mathcal{N}_k^M$, is given by the quantity \cite{dimiccoli2019enhancing}
\begin{equation}
\label{equ:NLsim}
S^{NL}({k,j}) = \frac{1}{\mathcal{Z}(k)} \exp \left(-\frac{ dist(X(\mathcal{N}_k^M), X(\mathcal{N}_j^M))} {h}\right).
\end{equation}
Here $\mathcal{Z}(k)$ is a normalizing factor such that $\sum_{j\in \mathcal{N}_k^L}S^{NL}_{kj}=1$, ensuring that $S^{NL}_{kj}$ can be interpreted as a conditional probability of $X(j)$ given $X(\mathcal{N}_k)$, as detailed in \cite{buades2005non}, $dist(X(\mathcal{N}_k),X(\mathcal{N}_j)) = \sum_{i=1}^{2M}   ||X(\mathcal{N}_k(i))-X(\mathcal{N}_j(i))||_{\ell_1}$ is the sum of the $\ell_1$ distances of the vectors in the neighborhoods of $k$ and $j$, and $h$ is the parameter that tunes the decay of the exponential function. The key idea of our graph initialization is to model each frame $k$ by its denoised version, obtained as
\begin{eqnarray}
\nonumber
\hat{X}(k) &=& \textnormal{NLmeans}^{1D}(X(k)) = {\sum}_{j \in \mathcal{N}_k^L} S^{NL}_{kj}\cdot X(j),\\
\hat X_0 &=&(\hat{X}(1)^T\; \hat{X}(2)^T\;\ldots\; \hat{X}(N)^T)^T.
\label{equ:NLmeans}
\end{eqnarray}
A numerical illustration on real data is provided in Figure~\ref{fig:graph_ini}.

\noindent{\bf Initial graph and initial embedding.\quad}
An initial graph $\mathcal{G}_0 $ is obtained by computing the $ \mathbb{R}^N \times \mathbb{R}^N$ affinity matrix $\textnormal{G}_0=S_{\hat l}(\hat{X}_0)$ of $\hat{X}_0$, defined elementwise as the pairwise similarity 
\begin{equation}
\label{equ:cosdist}
(S_l(X))_{kj} = \exp\left(-\frac{1-cdist(X(j),X(k))}{l}\right)
\end{equation}
where $cdist(\cdot,\cdot)$  is the cosine distance and $l$ the filtering parameter of the exponential function. In the following, we will not distinguish any longer between a graph $\mathcal{G}$ and the representation by its affinity matrix $\textnormal{G}$ and make use of both symbols synonymously. In our model, $\mathcal{G}_0$ represents the initial data structure in the original high dimensional space as a fully connected graph,  from which we aim to learn a graph in the embedding space that better encodes temporal and semantic constraints, denoted $\tilde{\mathcal{G}}$. To obtain an initial embedding $\tilde{X}_{0}$ for the graph, we apply PCA on $\hat{X}_0$, keep the $d$ major principal components $\tilde X$ and minimize the cross-entropy loss $\mathcal{L}$ between the affinity matrices $\textnormal{G}_0=S_{\hat l}(\hat{X}_0)$ and $S_{\tilde l}(\tilde{X}_0)$
\begin{equation}
\label{equ:init}
\tilde{X}_0=   {\argmin}_{\tilde{X}}\mathcal{L}(S_{\tilde l}(\tilde{X}),\textnormal{G}_0)
\end{equation}
where the different filtering parameters $\hat l$ and $\tilde l$ account for the different dimensionality of $\hat{X}_{0}$ and $\tilde{X}_{0}$. Even if PCA is a linear operator and  for  small  sets  of  high-dimensional  vectors dual  PCA  could be more appropriate \cite{giannakis2018topology}, we found it sufficient here for initializing the algorithm. The initial graph $\tilde{\mathcal{G}}_0$ in the embedding space is then given by $\tilde{\textnormal{G}}_0=S_{\tilde{l}}(\tilde{X}_0)$.

\subsection{DGE core alternating steps}
\label{subsec:DGEcore}

Given the initial embedding $\tilde{X}_{0}$ and graphs $\mathcal{{G}}_0$ and $\mathcal{\tilde{G}}_{0}$, the main loop of our DGE  alternates over the following two steps:
\begin{enumerate}
\item Assuming that $\mathcal{\tilde{G}}_{i-1}$ is fixed, update the node representations $\tilde{X}_i$.
\item Assuming that $\tilde{X}_i$ is fixed, update the graph $\mathcal{\tilde{G}}_{i}$.
\end{enumerate}
Step (1) is inspired from graph embedding methods, such as the ones reviewed in Section \ref{sec:geoLearn}, which have proved to be very good at encoding a given graph structure. Step (2) aims at enforcing temporal constraints and at fostering semantic similarity in the graph structure. 

\noindent{\bf Graph embedding update.\quad}
To estimate the graph embedding $\tilde{X}_i$ at iteration $i$ assuming that $\mathcal{\tilde{G}}_{i-1}$ is given, we solve 
\begin{equation}
    \label{eq:embedding}
\tilde{X}_i \!=\!   \argmin_{\tilde{X}} (1\!-\!\alpha) \mathcal{L}_1 (S_{\tilde l}(\tilde{X}),\tilde{\textnormal{G}}_{i-1})+\alpha\mathcal{L}_2 (S_{\tilde l}(\tilde{X}),\textnormal{G}_0).\!
\end{equation}
Here $\mathcal{L}_1$ and $\mathcal{L}_2$ are cross-entropy losses and $S_{\tilde{l}}(\cdot,\cdot)$ is the cosine-distance based similarity defined in \eqref{equ:cosdist}. The first loss term controls the fit of the representation $\tilde X$ with  the learnt graph $\tilde{\mathcal{G}}_i$ in low-dimensional embedding space. The second loss term quantifies the fit of the representation $\tilde X$ with the fixed initial graph ${\mathcal{G}}_0$ in high dimensional space and is reminiscent of shallow graph embedding.
The  regularization parameter $\alpha \in[0,1]$ controls the relative weight of each loss. Standard gradient descent can be used to solve \eqref{eq:embedding}. In the numerical experiments reported below, $150$ iterations of gradient descent with Barzilai-Borwein adaptive step size \cite{barzilai1988two} were used.

\noindent{\bf Graph structure update.\quad}
To obtain an estimate of the graph structure at the $i$-th iteration, say $\tilde{\mathcal{G}}_i$, assuming that $\tilde{X}_{i}$ is given, we start from an initial estimate for $\tilde{\mathcal{G}}_i$ as $\tilde{\textnormal{G}}_{i} = S_{\tilde l}(\tilde{X}_{i})$, and then make use of the model assumptions described in Section \ref{sec:formulation} to modify the graph: temporal adjacency, and semantic similarity. 

i) To foster similarity for temporally adjacent nodes, we apply two operations. First, local averaging of the edge weights defined as $\tilde{\textnormal{G}}_i \gets\tilde{\textnormal{G}}_{i} \ast \mathcal{K}_{p}$, where $\ast$ is the 2D convolution operator and $\mathcal{K}_{p}$ a $p \times p$ kernel that is here simply the normalized bump function. 
Second, application of the shrinkage operation
\begin{equation}
\label{equ:graph1}
(\tilde{\textnormal{G}}_i)_{kj} \gets (f_\eta(\tilde{\textnormal{G}}_i))_{kj} 
= \begin{cases} 
(1-\eta)(\tilde{\textnormal{G}}_i)_{kj} 
\;\;\text{if    } |k-j|\!>\!1\!\!\!\\ 
{\color{white}(1-\eta)}(\tilde{\textnormal{G}}_i)_{kj}  \;\;\text{otherwise,} 
\end{cases}%
\end{equation} 
which leaves the similarities of directly temporally adjacent nodes of $\mathcal{\tilde{G}}_i$ unchanged, but shrinks the weights of edges between nodes $k$ and $j$ that are not direct temporal neighbors by a factor $\eta$, $0<\eta < 1$, thus strengthens the temporal adjacency of the graph.

ii) To reinforce the semantic similarity of ${\tilde X}_i$, we first obtain a coarse estimate of the community structure of the graph $\mathcal{\tilde{G}}_i$. 
To this end, we apply a clustering algorithm on $\tilde{X}_{i}$, which yields estimated cluster labels $\mathcal{C}=(c_j)_{j=1}^{N_C}$, $c_j\in\{1, ...,{N_C}\}$, for each frame, that roughly correspond to semantic contexts, i.e., communities. Then we modify $\tilde{\mathcal{G}}_i$ using the non-linear operation defined by
\begin{equation}
\label{equ:graph2}
(\tilde{\textnormal{G}}_i)_{kj} \gets (g_\mu(\tilde{\textnormal{G}}_i, \mathcal{C}))_{kj} 
= \begin{cases} 
(1-\mu)(\tilde{\textnormal{G}}_i)_{kj} 
 \;\;\;\text{if    }  c_j\neq c_k\!\!
\\ 
{\color{white}(1-\mu)}(\tilde{\textnormal{G}}_i)_{kj}\;\;\;\text{otherwise.}\!\!
\end{cases}\!\!\!\!\!\!
\end{equation}
This graph update reduces the similarity between nodes $k$ and $j$ that do not belong to the same cluster, $c_j\neq c_k$, by a factor $\mu$, $0< \mu < 1$, and does not change similarities within clusters, hence reinforces within-event semantic similarity. 

Thus, both the embedding step and the clustering jointly contribute to learn representations that account for the semantics encoded by the initial CNN features.
DGE aims at revealing the temporal and semantic relatedness for each pair of data vectors, and therefore the estimated graphs $\mathcal{\tilde{G}}_i$, $i=0,\ldots,K$, are fully connected at each stage. 
A high-level overview of our DGE approach can be found on ALGORITHM \ref{alg:algo}.

\subsection{Graph post-processing: Event boundary detection}
Depending on the problem, applicative context and objective, different standard and graph signal processing tools can be applied to the estimated graph $\mathcal{\tilde{G}}_K$ in order to extract the desired information or to transform $\mathcal{\tilde{G}}_K$ \cite{wang2015global,ortega2018graph}. To evaluate the effectiveness of the learnt representation  $\tilde{X}_K$ corresponding to $\mathcal{\tilde{G}}_K$ for event segmentation, we use it as the input of a boundary detector.

\noindent{\bf Boundary detector.\quad}
To ensure a fair comparison of the learnt event representations with \cite{del2018predicting}, we use the same boundary detector as therein, with the same parameter values and thresholds. It is based on the idea that when a vector $\tilde X(k)$ representing frame $k$ corresponds to an event boundary, the distance between the predictions computed for it from past ($j<k$) and future ($j>k$) representation vectors is likely to be large. Consequently, \cite{del2018predicting} defines the boundary prediction function as the (cosine) distance between these contextual forward and backward predictions for frame $k$. Those frames for which the values of the boundary prediction function exceed a threshold are the detected event boundaries, see \cite{del2018predicting} for details.

Hereafter we call our temporal segmentation model relying on the features learnt by using the proposed DGE approach CES-DGE, in analogy with CES-VCP in \cite{del2018predicting} (where CES stands for contextual event segmentation and VCP for visual context prediction).

\subsection{Numerical illustration of CES-DGE on synthetic data}
\label{sec:synthexp}

To illustrate the main idea behind our modeling, we show with a synthetic example in $n=2$ dimensions how the original data $X$ and the associated initial graph $\mathcal{\tilde{G}}_0$ change over $K=10$ iterations of the DGE algorithm (here, $d=n=2$, and we assume $\tilde X_0=\hat X_0 = X$, i.e., no preprocessing). The data consist of a temporal sequence of $N=350$ feature vectors that are drawn from four different Gaussian distributions with mean vectors $(   -7.8\;    5.1),(-1.4\;    2.8),(-2.9\;   12.1),(2.5\;    3.4)$ and diagonal covariance matrix $\sigma \mathbf{I}$ with $\sigma=3.7$. The different distributions are selected according to a Markov switching model in which the probability to remain in a state decreases from $1$ at its onset at an exponential rate with time. This makes it likely that a reasonable number of temporally adjacent vectors are drawn from the same distribution, thus modeling a community corresponding with a semantic context.

Results are plotted in Figure~\ref{fig:synt} as scatter plots of the learnt representations $\tilde X_i$ (panel (a)) and as time series $\tilde X_i$ (panel (b)) for iterations $i=(0,1,3,10)$. Clearly, it is difficult to obtain a good segmentation for the initial, cleaned data (iteration $i=0$, left column). Yet, after a few DGE iterations the learnt representation vectors $\tilde X_i$ that  belong to the same context are aggregated at similar values (Figure~\ref{fig:synt} (a-b), columns 2 to 4, respectively). At the same time, those vectors that belong to different contexts are pushed away from each other in the representational space. This effectively increases the similarity within each semantic context.
An alternative view is provided by the corresponding learnt graphs $\mathcal{\tilde{G}}_i$ that are plotted in panels (c-d) in Figure~\ref{fig:synt}. It can be observed how increasingly homogeneous diagonal and off-diagonal blocks with clear boundaries emerge with progressing iteration number $i$, reflecting the temporal community structure underlying the data.
This improved representation leads in turn to significantly better event segmentation results, with F-score values increasing from initially $0.59$ to $0.62$, $0.64$, $0.67$ for iterations $i=1,3,4-10$, respectively (cf., Figure~\ref{fig:synt}, panel (e)).

\section{Performance evaluation}
\label{sec:exp} 

\subsection{Datasets and experimental setup}
\label{sec:exp:data}

\noindent{\bf Datasets.\quad}
We used two temporal segmentation benchmark datasets for performance evaluation. The EDUBSeg dataset, introduced in \cite{dimiccoli2017sr}, consists of 20 first-person image sequences acquired by seven people with a total of 18,735 images.  The dataset has been used to validate image sequence temporal segmentation in several recent works \cite{dimiccoli2017sr,dias2019learning,paci2016context,del2018predicting}. For each image sequence, three manual segmentations have been obtained by three different persons. In line with previous works, the first segmentation was used here as the ground truth. The average performance of the two remaining segmentations is used as an estimate of the manual segmentation performance. 
The second dataset is the larger and more recent EDUBSeg-Desc dataset \cite{bolanos2018egocentric}, with 46 sequences (42,947 images) acquired by eight people. 

On average, the sequences of both datasets contain roughly the same number of ground truth event boundaries (28 for the former vs.~26 for the latter). However, those of EDUBSeg-Desc consist of 25\% longer continuously recorded segments than EDUBSeg (3h46m25s vs.~3h1m29s continuous ``camera on'' time, and 3.0 vs.~3.55 continuously recorded segments per sequence). Since EDUBSeg-Desc is 50$\%$ larger than EDUBSeg and event transitions within the increased number of continuously recorded segments are more difficult to detect, EDUBSeg-Desc is considered more challenging \cite{bolanos2018egocentric}.

Other publicly available datasets of egocentric image sequences, such as CLEF \cite{dang2017overview}, NTCIR \cite{gurrin2017overview} and the more recent R3 \cite{del2018predicting}, do not have ground truth event segmentations. They can therefore not be used for performance evaluation. These datasets with more than 1.2 million images were used for training in \cite{del2018predicting}. We emphasize that in contrast, our algorithm operates without training dataset.

\noindent{\bf Feature extraction.\quad} As in \cite{del2018predicting}, each frame of the egocentric image sequences was described here using the output of the pre-pooling layer of InceptionV3 \cite{szegedy2016rethinking} pretrained on ImageNet, resulting in $n=2048$ raw features $X(k)$ per frame $k$.

\noindent{\bf Performance evaluation.\quad}
Following previous work  \cite{paci2016context,dimiccoli2017sr,del2018predicting,dias2019learning}, we consider a detected event boundary to be correct when it falls within a range of $\pm \tau$ around the position of a true boundary. We use F-score, Precision (Prec) and Recall (Rec) to evaluate the performance of our approach. While previous work considered only a single level of tolerance $\tau=5$, we here report results for several values $\tau\in (1,2,3,4,5)$.

\noindent{\bf DGE hyperparameters.\quad}
The hyperparameter values for the graph initialization and embedding (i.e., for the non-local self-similarity kernel \eqref{equ:NLsim} and for the similarity \eqref{equ:cosdist}) have been chosen a priori based on visual inspection of the similarity matrices of $\hat X_0$ and $\tilde{X}_0$ for a few sequences of EDUBSeg. The values are fixed to $L=3$, $M=1$, $h=0.25$ for Eq.~\eqref{equ:NLsim}, and $\hat l=0.0025$, $\tilde l=0.02 d$ for Eq.~\eqref{equ:cosdist}. The embedding dimension is set to $d=15$, which is found sufficient for the representation to faithfully reproduce the graph topology underlying the data. The influence of $d$ is reported in the next section. The DGE core loop hyperparameters are set to $K=2$ (DGE iterations), $\alpha = 0.1$ (graph embedding update), $p=3$, $\eta=0.3$ (temporal prior), and $\mu=0.1$, $N_C=10$ (semantic prior; a k-means algorithm is used to estimate cluster labels). 
These hyperparameter values have been selected by a grid search using the EDUB-Seg dataset.
Our grid search strategy consisted in first tuning the embedding dimension $d$ based on the quality of the initial embedding, and to perform preliminary individual line searches to determine search ranges and granularity for the remaining hyperparameters.
The values retained by the grid search for the EDUB-Seg dataset are also used for the EDUB-SegDesc dataset, without modification; robustness of these choices is also investigated in the next section. 

\subsection{Robustness to changes in hyperparameter values}
\label{subsec:hyperprameters}

Table~\ref{tab:robustness} reports F-score values obtained on the EDUBSeg dataset when the embedding dimension $d$, the DGE iterations $K$ and the DGE core parameters $\alpha$, $p$, $\eta$, $\mu$ and $N_{C}$ are varied each individually. It can be appreciated that the performance of CES-DGE is overall robust w.r.t. precise hyperparameter values. 
As long as the embedding dimension $d$ is chosen not too small and not too large, F-score values vary little (no more than 3 percentage points below the best observed F-score values for the range of embedding dimensions $10<d<50$) ; this corroborates similar findings on the existence of a trade-off between low-dimensionality and fidelity to the graph structure for node embedding in different contexts, cf., e.g., \cite{hamilton2017representation,ahmed2013distributed,ou2016asymmetric,cao2015grarep,perozzi2014deepwalk,grover2016node2vec,tang2015line,chen2018harp,cao2016deep,wang2016structural,hinton2006reducing,hamilton2017inductive}.
The highest sensitivity to DGE core loop hyperparameters is observed for the DGE iteration number $K$, whose optimal value is a compromise between increasing the similarity  (i.e., choosing $K$ large) between $\tilde X_i$ within correctly detected communities but not increasing it too much (i.e., $K$ small) within incorrectly alienated communities.
If chosen as $1<K<5$, F-score values are at most 3 percentage points below the best observed F-score.
Results are also very robust w.r.t. temporal regularization for reasonably small parameter values of $p\leq 5$ and $\eta\leq 0.4$. For larger values the learnt representation is over-smoothed. Similarly, F-score values vary little when changing the semantic similarity parameter as long as $0<\mu\leq 0.2$. Note that these variations are significant because $\eta,\mu\in(0,1)$. Finally, F-score values drop by less than 3 percentage points when the number of clusters 
is selected within a reasonably large range $N_{C}\in(6,\ldots,20)$. Overall, these results suggest that CES-DGE is quite insensitive to hyperparameter tuning and yields robust segmentation results for a wide range of hyperparameter values.

\setlength{\tabcolsep}{2pt}
\begin{table}[]
\centering
\begin{tabular}{ c | c c c c c}
\hline
Parameter & \multicolumn{5}{c}{\textbf{F-score}}\\ 
\hline 
\multicolumn{5}{c}{\scriptsize\vspace{-1.7mm}}\\
\hline
$d$ & 5 &7 &10 &15 &20    \vspace{-0.5mm} \\
&0.59& 0.61& 0.65& \textbf{0.70} & 0.69 \\
\scriptsize(embedding dimension)  &25 &35 &45 &100 &150   \vspace{-0.5mm}\\
& 0.68& 0.67& 0.68& 0.65& 0.64 \\\hline
\multicolumn{5}{c}{\scriptsize\vspace{-1.7mm}}\\
\hline
$K$ &1&2&3&4&5  \vspace{-0.5mm}\\
\scriptsize(DGE iterations)&  0.66& \textbf{0.70}& 0.68& 0.67& 0.66 
\\\hline
\multicolumn{5}{c}{\scriptsize\vspace{-1.7mm}}\\
\hline  
$\alpha$  &0.05 &0.1 &0.2 &0.3 &0.4   \vspace{-0.5mm}\\
\scriptsize(graph embedding update)&0.679& \textbf{0.70}& 0.69& 0.69& 0.68 \\\hline
$p$  & 2 &3 &4 &5 &6   \vspace{-0.5mm}\\
\scriptsize(2D local average size)&0.69 &\textbf{0.70} &0.69 &0.69 &0.68 \\\hline
$\eta$  & 0.01& 0.1& 0.3& 0.4& 0.6    \vspace{-0.5mm}\\
\scriptsize(graph temporal regularization)&0.69&  0.69& \textbf{0.70}& 0.69& 0.67 \\\hline
$\mu$ &  0.03& 0.05& 0.1& 0.2& 0.3   \vspace{-0.5mm}\\
\scriptsize(extra-cluster penalty)&  0.68& 0.70&  \textbf{0.70}&  0.69&  0.67 \\\hline
$N_{C}$ & 3 &4 &6 &8 &10      \vspace{-0.5mm}\\
&0.65& 0.66& 0.67& 0.68& \textbf{0.70}  \\
\scriptsize(cluster number)  &12 &14 &16 &18 &20   \vspace{-0.5mm}\\
& 0.69& 0.69& 0.68& 0.67& 0.68 \\\hline
\end{tabular}
\caption{Robustness of CES-DGE with respect to hyperparameter values (EDUBSeg, tolerance $\tau=5$): F-scores obtained when varying the DGE hyperparameter indicated in the first column, with all others held fixed (best results in bold).}
\label{tab:robustness}
\end{table}

\subsection{Comparative results for EDUBSeg dataset}
\label{subsec:edub}

\setlength{\tabcolsep}{6pt}
\begin{table}[tb]
\centering
\begin{tabular}{ c c c c}
\hline
Method & \textbf{F-score} & \textbf{Rec} & \textbf{Prec} \\ \hline
 \hline
k-Means smoothed & 0.51 & 0.39 & 0.82\\
\hline 
AC-color & 0.38  & 0.25  &  0.90\\
\hline 
SR-ClusteringCNN & 0.53 & 0.68 & 0.49 \\
\hline 
KTS & 0.53 & 0.40 & 0.87 \\
\hline 
CES-TSC& 0.66 & 0.69 & 0.67\\
\hline 
CES-VCP & 0.69 & \textbf{0.77} & 0.66  \\
\hline 
{CES-DGE}&  \textbf{0.70} & 0.70 & \textbf{0.72}\\
\hline\hline
Manual segmentation & \textbf{0.72} & 0.68 & \textbf{0.80}\\
\hline 
\end{tabular}
\caption{Comparison of CES-DGE with state-of-the-art methods \& manual segmentation for EDUBSeg and tolerance $\tau=5$.}
\label{tab:compALL}
\end{table}

\setlength{\tabcolsep}{2pt}
\begin{table}[t]
\centering
\begin{tabular}{ c | c c c | c c c}
\hline
\multicolumn{1}{c}{Method} & \multicolumn{3}{c}{CES-VCP}& \multicolumn{3}{c}{CES-DGE}\\
\hline
Tolerance & \textbf{F-score} & \textbf{Rec} & \textbf{Prec} & \textbf{F-score} & \textbf{Rec} & \textbf{Prec} \\ \hline\hline
$\tau=5$ & 0.69 & \textbf{0.77} & 0.66  &\textbf{0.70} & 0.70 & \textbf{0.72}  \\ \hline 
$\tau=4$ & 0.67 & \textbf{0.75} & 0.63 &  \textbf{0.68} & 0.69 & \textbf{0.70}  \\ \hline 
$\tau=3$ &  0.64 & 0.62 & \textbf{0.71}  &  \textbf{0.65} & \textbf{0.64} & {0.68}  \\ \hline 
$\tau=2$ & \textbf{0.59} & \textbf{0.67} & 0.56  &  \textbf{0.59} & 0.59 & \textbf{0.61}  \\ \hline 
$\tau=1$ & {0.44} & {0.44} & 0.49 &  \textbf{0.48} & \textbf{0.48} & \textbf{0.50}  \\
\hline 
\end{tabular}
\caption{Comparison of CES-DGE with state-of-the-art CES-VCP for different values of tolerance for EDUBSeg.}
\label{tab:compCES}
\end{table}

\setlength{\tabcolsep}{2pt}
\begin{figure*}[tb]
   \centering
\includegraphics[width=0.98\linewidth]{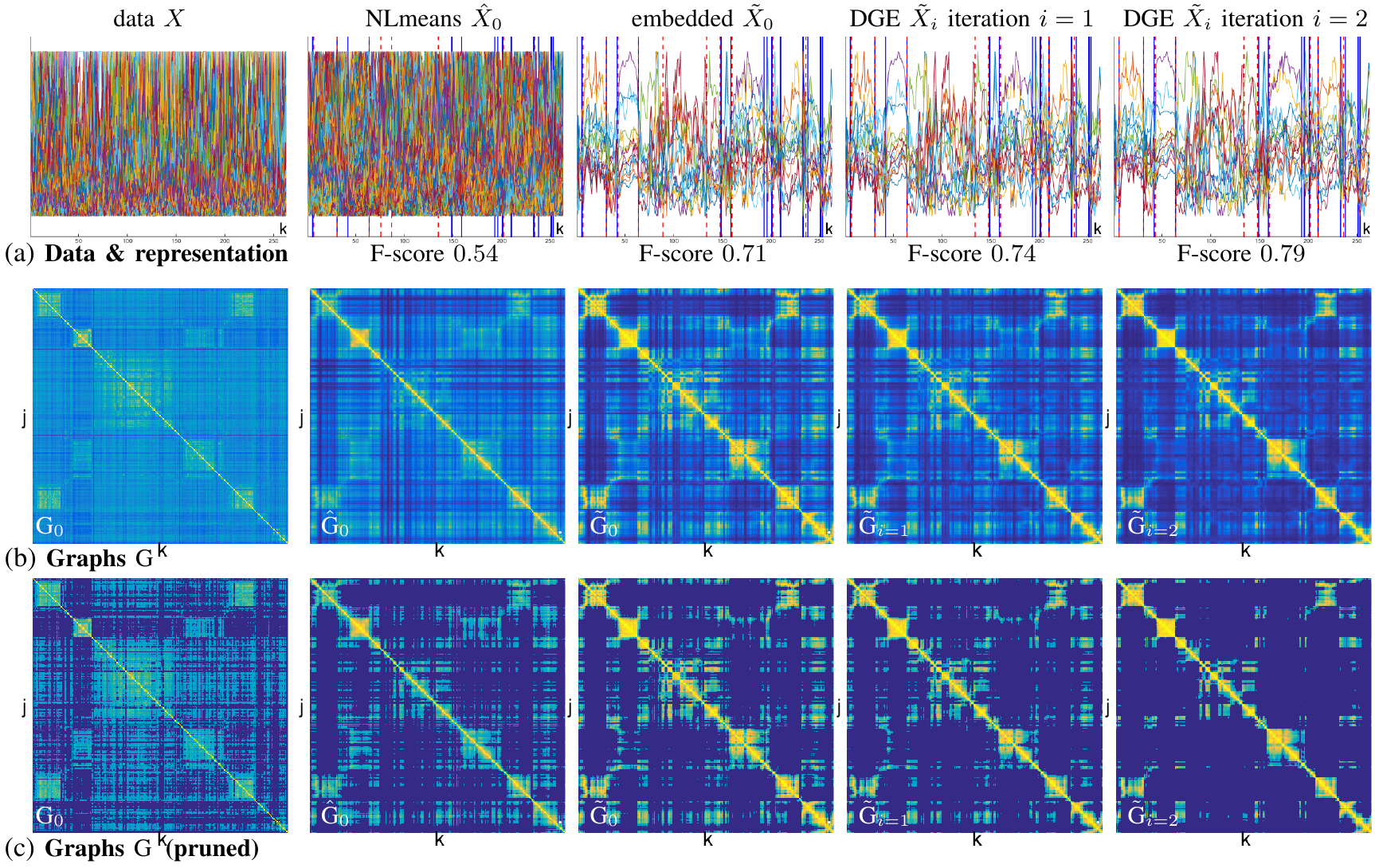}\vspace{-3mm}
\caption[]{%
{\bf Illustration of DEG on Subject 1 Day 3 of EDUBSeg.} Panel (a): Data and learnt representations with boundary estimates (red dashed vertical bars), ground truth (blue solid vertical bars) and resulting F-score values; from left to right initial features $X$ (1st column), denoised features $\hat X_0$ (2nd column), initial embedded features $\tilde X_{i=0}$ (3rd column), learnt representation $\tilde X_{i}$ at iterations $i=1$ and  $i=2$ (4th \& 5th column, respectively). Panel (b) plots the corresponding graphs ${\mathcal{G}}_0$, $\hat{\mathcal{G}}_0$ and $\tilde{\mathcal{G}}_i$ (from left to right). Panel (c) shows the graph after edges with weight smaller than 0.3 times the strongest edge have been removed (dark blue corresponds with small weights and yellow corresponds with large weights, respectively).
}
\label{fig:graph_ini}
\end{figure*}

Table~\ref{tab:compALL} reports comparisons with five state-of-the-art methods for a fixed value of tolerance ($\tau=5$, results reproduced from \cite{del2018predicting}). The first four (k-means smoothed with k$\;=30$, AC-color \cite{lee2015predicting}, SR-Clustering \cite{dimiccoli2017sr}, KTS \cite{potapov2014category}) are standard/generic approaches and achieve modest performance, with F-scores no better than 0.53.
CES-VCP of \cite{del2018predicting} yields significantly better F-score of 0.69, thanks to the use of a large training set for learning the event representation. The proposed CES-DGE approach further improves on this state-of-the-art result and yields F-score of 0.70. Interestingly, CES-DGE also achieves more balanced Rec and Prec values of 0.70 and 0.72, as compared to 0.77 and 0.66 for CES-VCP. Moreover, even when compared to average manual segmentation performance our results are only 2 percentage points below.
Finally, we report results obtained by combining the Temporal Subspace Clustering (TSC) algorithm of \cite{li2015temporal} with the CES segmentation of \cite{del2018predicting} (CES-TSC), yielding F-score values 4 percentage points below the proposed CES-DGE.

In Table \ref{tab:compCES}, we provide comparisons between our proposed approach and CES-VCP for different values of tolerance. We can observe that CES-DGE achieves systematically better results than CES-VCP in terms of F-score for all values of tolerance. These improvements with respect to the state of the art reach up to 4 percentage points. Besides, Rec/Prec values for CES-DGE are more balanced and within $\pm$3 percentage points of the values for F-score for all tolerance levels ($\pm$8 percentage points of F-score for CES-VCP).

Overall, this leads  to conclude that the proposed CES-DGE approach is effective in learning event representations for image sequences and yields robust segmentation results. These results are even more remarkable considering that CES-DGE learns the feature representation from the image sequence itself, without relying on any training dataset.

\subsection{Comparative results for EDUBSeg-Desc dataset}
\label{subsec:edubdesc}
Table~\ref{tab:compCESdesc} summarizes the event boundary detection performance of the proposed CES-DGE approach and of CES-VCP for the larger EDUBSeg-Desc dataset. Since CES-VCP reported state-of-the-art results with F-score values 16 percentage points above other methods, cf., Table~\ref{tab:compALL}, we omit comparison with other methods in what follows for space reasons. The same (hyper)parameter values as for EDUBSeg are used, without any modification. It can be observed that the performance for both CES-VCP and CES-DGE are inferior to those reported for the EDUBSeg dataset in the previous section, for all tolerance values. These results corroborate that EDUBSeg-Desc contains more difficult image sequences than EDUBSeg (see Section \ref{sec:exp:data} and \cite{bolanos2018egocentric}). Interestingly, while the F-scores achieved by CES-VCP are up to 12 percentage points (and more than 11 percentage points on average) below that reported for EDUBSeg, the F-scores of the proposed CES-DGE approach are at worst 5 percentage points smaller. In other words, CES-DGE yields up to 8 percentage points (on average 6 percentage points) better F-score values than the state of the art for the EDUBSeg-Desc dataset. Our CES-DGE also yields systematically better Rec and Prec values, for all levels of tolerance. Overall, these findings corroborate those obtained for the EDUBSeg dataset and confirm the excellent practical performance of the proposed approach. In particular, the results suggest that the proposed approach effectively avoids domain adaptation problems since it does not rely on a training dataset. 
It is interesting to note that CES-TSC also yields quite robust results (F-score/Rec/Prec values 0.63/0.60/0.70 for $\tau=5$) for similar reasons, yet with performance significantly below our CES-DGE.

\setlength{\tabcolsep}{2pt}
\begin{table}[tb]
\centering
\begin{tabular}{ c | c c c | c c c}
\hline
\multicolumn{1}{c}{Method} & \multicolumn{3}{c}{CES-VCP}& \multicolumn{3}{c}{CES-DGE}\\
\hline
Tolerance & \textbf{F-score} & \textbf{Rec} & \textbf{Prec} & \textbf{F-score} & \textbf{Rec} & \textbf{Prec} \\ \hline\hline
$\tau=5$ & 0.57 & {0.59} & 0.60  & \textbf{0.65} & \textbf{0.67} & \textbf{0.65}  \\ \hline 
$\tau=4$ & 0.56 & {0.58} & 0.58  &  \textbf{0.63} & \textbf{0.66} & \textbf{0.63}  \\ \hline 
$\tau=3$ & 0.52 & 0.54 & {0.54}  &  \textbf{0.60} & \textbf{0.62} & \textbf{0.60}  \\ \hline 
$\tau=2$ & {0.49} & {0.50} & 0.50 &  \textbf{0.54} & \textbf{0.56} & \textbf{0.54} \\ \hline 
$\tau=1$ & {0.43} & {0.44} & \textbf{0.45} &  \textbf{0.45} & \textbf{0.46} & \textbf{0.45}  \\
\hline 
\end{tabular}
\caption{Comparison of CES-DGE with state-of-the-art CES-VCP for different values of tolerance for EDUBSeg-Desc.}
\label{tab:compCESdesc}
\end{table}

\subsection{Ablation study}
\label{subsec:robust}

\noindent{\bf Graph initialization.\quad}
We investigate performance obtained by applying the boundary detector to the features obtained at different stages of our method. First, the original features $X$ (denoted CES-raw) and the features $\hat X_0$ obtained by applying NLmeans on the temporal dimension (denoted CES-NLmeans-1D), both  of dimension $n=2048$. Second, the initial embedded features $\tilde{X}_0$ (denoted CES-Embedding) and the features obtained after running the DGE main loop for $K=2$ iterations (denoted CES-DGE), both of dimension $d=15$. The results obtained for the EDUBSeg dataset are reported in Table~\ref{tab:Ablation}. They indicate that CES-NLmeans-1D increases F-score by 8 percentage points w.r.t. CES-raw, and CES-Embedding adds another 1 percentage points in F-score. This confirms that the graph initialization and the reduction of the dimension of the graph representation is beneficial. CES-DGE gains an additional 9 percentage points in F-score value, hence significantly improves upon this initial embedding. 
An illustration of the effect of the graph initialization and of the DGE steps for EDUBSeg Subject 1 Day 3 is provided in Figure~\ref{fig:graph_ini}. It can be observed how the boundaries between temporally adjacent frames along the diagonal in the graph are successively enhanced as the original features $X$ (column 1) are first replaced with the denoised version $\hat{X}_0$ (column 2), then with the embedded features $\tilde X_0$ (column 3), and finally with the DGE representation estimates (columns 4 \& 5 for DGE iterations $i=1,2$, respectively). Moreover, the boundaries of off-diagonal blocks, which indicate frames at different temporal locations that presumably belong to the same community, are sharpened.

\noindent{\bf  DGE core operations.\quad}
In Table~\ref{tab:Ablationcore}, we report the performance that is obtained on the EDUBSeg dataset when the different operations in the DGE core iterations are removed one-by-one by setting the respective parameter to zero: graph embedding update regularization ($\alpha$), edge local averaging ($p$),  temporal edge weighting ($\eta$), and extra-cluster penalization ($\mu$). It is observed that the overall DGE F-score drops by 1 to 3 percentage points when one single of these operations is deactivated (versus a drop of 9 percentage points from 0.70 to 0.61 when no DGE operation is performed at all, as discussed in the previous paragraph). The fact that removing any of the operations individually does not lead to a knock-out of the DGE loop suggests that the associated individual (temporal \& semantic) model assumptions are all and independently important. Among all operations, the largest individual F-score drop (3 percentage points) corresponds with deactivating the extra-cluster penalization (i.e., $\mu=0$). This points to the essential role of semantic similarity in the graph model. The graph temporal edge weighting is also effective for encoding the temporal prior (2 percentage points F-score drop if deactivated, i.e., $\eta=0$). The smallest F-score difference (1 percentage point) is associated with edge local averaging (i.e., $p=0$). To improve this additional temporal regularization step, future work could use nonlocal instead of local averaging, or learnt kernels.

\setlength{\tabcolsep}{6pt}
\begin{table}[tb]
\centering
\begin{tabular}{ c c c c}
\hline
Method & \textbf{F-score} & \textbf{Rec} & \textbf{Prec} \\ \hline
 \hline
CES-raw & 0.52 & 0.56 & 0.56  \\ \hline 
CES-NLmeans-1D & 0.60 & 0.63 & 0.61  \\ \hline 
CES-Embedding & 0.61 & 0.61 & 0.65  \\ \hline 
{CES-DGE}&  0.70 & 0.70 & 0.72\\
\hline 
\end{tabular}
\caption{CES-DGE ablation study for EDUBSeg and tolerance $\tau=5$.}
\label{tab:Ablation}
\end{table}

\setlength{\tabcolsep}{2.7pt}
\begin{table}[]
\centering
\begin{tabular}{ c | c c c c}
\hline
Deactivated DGE parameter &  $\alpha=0$ & $p=0$ & $\eta=0$ & $\mu=0$  \\ \hline
\textbf{F-score} & \textcolor{white}{$-$}0.69 & \textcolor{white}{$-$}0.69 & \textcolor{white}{$-$}0.68 & \textcolor{white}{$-$}0.67 \\ \hline
difference with full DGE (0.70) & $-$0.01 & $-$0.01 & $-$0.02 & $-$0.03 \\ \hline
\end{tabular}
\caption{F-scores obtained when single core steps  are removed from DGE (indicated by a zero value for the respective parameter).%
}%
\label{tab:Ablationcore}
\end{table}

\subsection{Generalization capabilities}
\label{subsec:generalization}

{\noindent{\bf Human Motion Segmentation.\quad}
To study the generalization capabilities of DGE to learn representations suitable for temporal segmentation beyond first-person image sequences, we applied it to two well established benchmark datasets for Human Motion Segmentation (HMS),
Keck \cite{lin2009recognizing} and MAD \cite{huang2014sequential}.
These datasets  consist of 3rd person shots
with static background
of a single person acting short motions, captured at high temporal resolution. Here, events correspond with typical motions (e.g., jumping, walking), and our approach models frames corresponding with a sequence of body poses of a motion (e.g., squad+standing+squad+\ldots) as semantically related.
The Keck and MAD datasets have been chosen among the four HMS benchmarks used by the state-of-the-art works because they are considered particularly challenging due to variable background (Keck) and large number of motions and subjects (MAD), respectively \cite{wang2018low,zhou2020multi}.
The state of the art for HMS has been reported in \cite{wang2018low} using a Low Rank Transfer subspace (LRT) model, and very recently in \cite{zhou2020multi} using a Multi-mutual transfer subspace learning (MTS) model.
We use the same HOG features as therein\footnote{https://github.com/wanglichenxj/Low-Rank-Transfer-Human-Motion-Segmentation}.
Unlike \cite{wang2018low} and \cite{zhou2020multi}, we use the same set of parameters for both datasets.
Moreover, with respect to the previous sections, we only updated the embedding dimension and temporal prior (to $d=35$,  and $p=50$, using grid search) and maintain the same values as above for the remaining hyperparameters ($K=2$, $N_C=10$, $\alpha=0.1$, $\eta=0.3$, $\mu=0.1$). Results are reported in Table~\ref{tab:HMS} in terms of Clustering Accuracy (ACC) and Normalized Mutual Information (NMI). Our DGE achieves state-of-the-art performance also for this different problem. This shows that it is a general framework that can be used in other contexts.}

\noindent{\bf Computational cost.\quad}
DGE has complexity $\OO{N^2}$ since all of the operations in Algorithm 1 have at most that complexity. To give a concrete example of execution time, it required $\sim$2 minutes on a standard Dell Precision 7920 Tower with one NVIDIA Titan XP to process the EDUB-Seg dataset (20 sequences with a total of $18,735$ frames).

\setlength{\tabcolsep}{3.7pt}
\begin{table}[]
\centering
\begin{tabular}{ l | c c | c c}
\hline
Dataset& \multicolumn{2}{c|}{Keck}& \multicolumn{2}{c}{MAD}\\ \hline\hline
Method& \bf ACC & \bf NMI & \bf ACC & \bf NMI \\ \hline
TSC  & 0.48 & 0.71 & 0.56 & 0.77 \\ \hline
LRT  & 0.55 & 0.82 & 0.60& 0.82 \\ \hline
MTS  & 0.60 & \bf0.83 & 0.62 & \bf0.83 \\ \hline
DGE &\bf 0.72 & \bf0.83 & \bf 0.67 & 0.82 \\ \hline 
\end{tabular}
\caption{Comparison with state of the art for Human Motion Segmentation for the Keck and MAD benchmark datasets (baselines taken from \cite{wang2018low} and \cite{zhou2020multi}).%
}%
\label{tab:HMS}
\end{table}

\section{Conclusion}
\label{sec:conclusions}
This paper proposed a novel approach to learn representations of events in low temporal resolution image sequences, named Dynamic Graph Embedding (DGE). Unlike state-of-the-art work, which requires (large) datasets for training the model, DGE operates without any training set and learns the temporal event representation for an image sequence directly from the sequence itself.To this end, we introduced an original model based on the assumption that the sequence can be represented as a graph that captures both the temporal and the semantic similarity of the images, which is understood here as similarity in terms of high-level visual features. The key novelty of our DGE approach is then to learn the structure of this unknown underlying graph jointly with a low-dimensional graph embedding. Experimental results have shown that DGE yields robust and effective event representations for temporal segmentation. It outperforms the state of the art in terms of event boundary detection precision, improving F-score values by 1 and 8 percentage points on the EDUBSeg and EDUBSeg-Desc event segmentation benchmark datasets, respectively. 
Moreover, we showed the generalization capabilities of the proposed DGE to the problem of Human Motion Segmentation.
Future work will include exploring the use of more sophisticated methods than the k-means algorithm in the semantic similarity estimation step, and the study of extensions and applications in the field of video analysis such as video and motion segmentation, action detection, and action proposal generation.

\bibliographystyle{IEEEtran}

\end{document}